%% file: hon_white_paper.tex
\documentclass{article} 
\usepackage{hon_white_paper,times,listings,subcaption, xcolor,graphicx,tcolorbox,booktabs,multirow}
\usepackage[subrefformat=parens]{subcaption}

\input{math_commands.tex}

\newcommand{\hon}{{\textit{“Human or Not?”}}}

\usepackage{hyperref}
\usepackage{url}

\title{Human or Not? \includegraphics[width=0.2\linewidth]{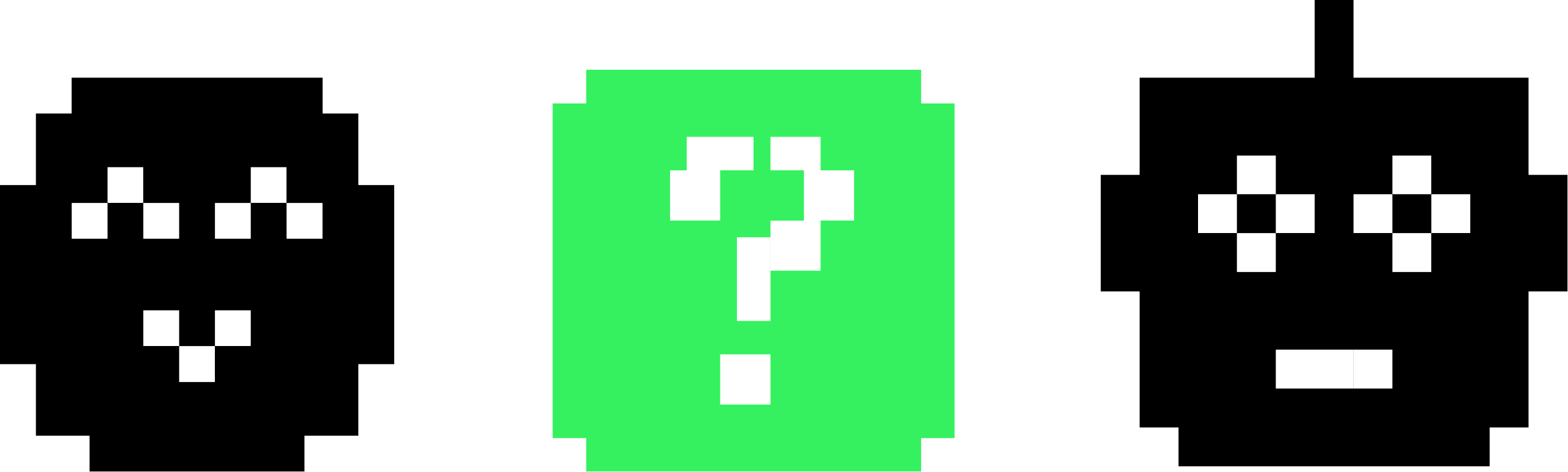}\\ \large A Gamified Approach to the Turing Test}

\author{Daniel Jannai \\
AI21 Labs \\
\texttt{danielj@ai21.com}
\And
Amos Meron \\
AI21 Labs \\
\texttt{amosm@ai21.com}
\And
Barak Lenz \\
AI21 Labs \\
\texttt{barakl@ai21.com}
\And
Yoav Levine \\
AI21 Labs \\
\texttt{yoavl@ai21.com}
\And
Yoav Shoham \\
AI21 Labs \\
\texttt{yoavs@ai21.com}
}

\usepackage{todonotes}

\begin{document}

\maketitle

\textit{“I believe that in 50 years’ time it will be possible to make computers play the imitation game so well that an average interrogator will have no more than 70\% chance of making the right identification after 5 minutes of questioning.”
}

-- Alan Turing, 1950

\begin{abstract}

We present \hon\footnote{\url{https://www.humanornot.ai/}}, an online game inspired by the Turing test, that measures the capability of AI chatbots to mimic humans in dialog, and of humans to tell bots from other humans. Over the course of a month, the game was played by over 1.5 million users who engaged in anonymous two-minute chat sessions with either another human or an AI language model which was prompted to behave like humans. The task of the players was to correctly guess whether they spoke to a person or to an AI. This largest scale Turing-style test conducted to date revealed some interesting facts. For example, overall users guessed the identity of their partners correctly in only 68\% of the games. In the subset of the games in which users faced an AI bot, users had even lower correct guess rates of 60\% (that is, not much higher than chance). This white paper details the development, deployment, and results of this unique experiment. While this experiment calls for many extensions and refinements, these findings already begin to shed light on the inevitable near future which will commingle humans and AI.

\end{abstract}

\section{Introduction}

The famous Turing test, originally proposed by Alan Turing in 1950 as “the imitation game”~\citep{turingtest}, was proposed as an operational test of intelligence, namely, testing a machine's ability to exhibit behavior indistinguishable from that of a human. In this proposed test, a human evaluator engages in a natural language conversation with both another human and a machine, and tries to distinguish between them. If the evaluator is unable to tell which is which, the machine is said to have passed the test.

While when it was proposed by Turing the test was more of a thought experiment than a practical proposal, in 1990, the Loebner Prize was established as an annual competition to reward the most human-like computer programs, adding a tangible goal of $100,000$\$ for the builders of an AI system that can fool all $4$ human judges. A widely publicized case of an AI system purportedly passing a Turing-like test emerged in 2014. Eugene Goostman, a chatbot emulating a 13-year-old Ukrainian boy, managed to convince 33\% of the judges at a competition held in the Royal Society in London that it was human. However, some argued that Goostman's portrayal as a young non-native English speaker was deliberately used to elicit forgiveness from those interacting with him, explaining any grammatical errors or gaps in general knowledge.

Since then, staggering progress has been made in the fields of artificial intelligence and natural language processing by Large language models (LLMs) like ChatGPT~\citep{chatgpt} or AI21 Labs' Jurassic-2~\citep{jurassic2}. Contemporary LLMs demonstrate remarkable language generation capabilities, producing coherent and contextually relevant responses across a wide range of topics. Indeed, while it is unlikely that Turing himself could have predicted the recent burst of AI advances, it is now clear that LLMs can be put to Turing-like tests with a fighting chance.

This white paper describes \hon, a social experiment that we released as a game in which users conduct open ended short conversations with a second party, and at the end cast their vote: did they converse with a fellow human user or with an AI bot?

The experiment was deliberately open-ended. While the explicit task given was to guess the type of interlocutor, users were free to add other motivations. Thus, some users tried to trick their partners into believing they are speaking with an AI, some tried to convince the other party that they are humans, while some users stuck to the assigned task and focused on interrogating their partner on what they considered to be traits or topics that distinguish between humans and bots. It should also be said that our AI bots too were not innocuous; we prompted them to make convincing attempts to mimic humans in a variety of aspects, which ranged from human-like slang and spelling errors, to holding a coherent back story about their character, all the way to leaving the game in the middle if the other side offended them. These made the game challenging and engaging, extracting emotional reactions from users at times.

Riding the current massive wave of public interest in AI, in its first month \hon~ accrued over 10 million human-AI and human-human conversations by over 1.5 million unique users, providing us with the first ever statistically robust scores to a Turing-like test. Several interesting findings emerged. Most importantly, our experiment echoed Turing's prediction that after a short interaction, an average interrogator would have less than 70\% chance of identifying an AI: users guessed the identity of their partners correctly in 68\% of the games (notably Turing assumed 5-minute interactions while we only allowed 2-minute ones). Intriguingly, in the subset of the games where users faced an AI bot, users had even lower correct guess rates of 60\%. While this isn't a completely fair comparison due to the shorter time frame and potential influence from game design decisions, it's fascinating to see Turing's forecast partially borne out. Although contemporary AI bots are still far from perfect, the results of our experiment clearly show that they are making staggering progress. The \hon~ setup is the first statistically robust method for tracking this progress, and it can be re-used in upcoming years as AI agents improve. Future analyses of this data can offer valuable insights into the current capabilities of AI models and the strategies humans use to identify AI-generated text.

Below, we outline the design and development process of \hon~ and present an initial analysis of the game's data. We hope that our setup and findings can provide valuable insights for the ongoing development of AI language models, the design of future human-AI interaction scenarios, and our understanding of how humans perceive and interact with AI systems.

\begin{table}[htbp]
  \centering
  \begin{tabular}{lccc}
    \toprule
    & Probability of Correct Guess \\
    \midrule
    Overall & 68\% \\
    When Partner is a Bot & 60\% \\
    When Partner is Human & 73\% \\
    \bottomrule
  \end{tabular}
  \caption{Probability of correct guess by partner type.}
\end{table}

\section{Game Design and Development}

\subsection{Motivation and Design Principles}

Contemporary AI models give us a glimpse into a future where AI plays active roles in our lives, ranging from providing chatbot assistance in commercial services, revolutionizing education, boosting creativity as a thought partner for creators, providing loneliness relief for the elderly, and more. Given this trajectory, we think it is important to (1) understand the traits and behaviors which people perceive as “human-like” or “machine-like”, and (2) develop quantitative measures that capture the ability of AI systems to mimic humans. With this in mind, we created a platform that would facilitate Turing-like tests in a modern, engaging, and accessible manner. Our success in popularizing this experiment provides the first ever statistically robust score to a Turing-like test, which serves as a baseline for future progress.

Concretely, we made strategic choices aimed at creating an immersive gamified experience which encourages recurring users. The conversations have a “ping-pong” structure that prevents players from sending two consecutive messages without a response, in order to ensure a balanced and dynamic exchange. Each message, limited to a maximum of 100 characters, has to be composed and sent within a 20-second window, and the chat ends after 2 minutes, usually consisting of 4-5 messages from each side. This ensures that players don't have to wait for too long, so they can remain engaged with the game and a constant suspense is kept. Once the conversation is over, players are prompted to guess whether their conversational partner was a fellow human or an AI bot.

Several other design decisions shaped the game dynamics. Firstly, input was limited to Latin characters and emojis to encourage English communication, though this solution was only partially effective as many languages can still be written using Latin characters. Secondly, we opted for anonymity, not requiring any registration, which aimed to lower barriers to entry, although it limited demographic analysis. In addition, we did not impose a limit on the number of times a user could play, providing the opportunity for them to develop and refine their strategies over time. Lastly, we refrained from implementing a leaderboard to keep the focus on exploring AI-human interaction and discourage system gaming. The only performance indicator was the display of correct guesses versus total games played.

Notably, we decided not to inform the players on what their counterpart's guess was (when it was human). The rationale behind this choice was to prevent incentivizing the player to imitate a bot. While our results showed that bot imitation was indeed a prevalent strategy used by the players, we suspect that the situation could have been exacerbated if players had access to their counterpart's eventual guess. As we reflect on the game design and user feedback, it is intriguing to consider alternative structures that could lead to different behaviors. For instance, one idea is a modified ranking system that penalizes users for being misidentified as bots, thus encouraging “authentic” human-like behavior. Such changes might further reduce bot imitation but could introduce new biases and strategies. It also raises intriguing questions about what constitutes “authentic” behavior in such a setting and how it might be incentivized.

In addition to what was previously mentioned, each message goes through a moderation service to ensure a safe environment and prevent abuse and hate speech. Any flagged content in AI-generated responses is filtered out, and if a user message is flagged, the conversation promptly ends. Finally, to encourage engaging and varied conversations, we provide both human users and AI bots with randomized conversation starters. These suggestions are intended to reduce the likelihood of repetitive or mundane conversations, contributing to the game's challenge and entertainment value.

\subsection{Developing the Conversational AI Bots}

One of the central challenges we faced during the development was ensuring that our AI bots were not easily distinguishable from humans. We recognized the difficulty of rendering bots truly human-like and hence set about defining a set of properties that the AI could emulate.

Consequently, we moved beyond Turing's original formulation which implicitly assumed the AI to be a neutral entity. Instead, we created a diverse array of bots, each with its unique personality and objective. We were motivated by the desire to keep the conversations interesting and less repetitive for recurring users, and to obscure the tell-tale signs of bots, making detection more challenging.

Specifically, each bot is prompted with a persona that includes basic information such as a name, age, occupation, and location, along with distinctive personality traits like wit, humor, or seriousness (see example in figure \ref{fig:simple_prompt}). The prompts also contained game instructions, making the bots aware of the gameplay context. To circumvent users leveraging the multilingual training data of the models to differentiate them from humans, the bots were also instructed to role-play a character that only spoke English. Some bots are even endowed with playful objectives, creating captivating narratives that keep the users engaged (see examples in figures \ref{fig:funny_prompt_1} and \ref{fig:funny_prompt_2}).

\begin{figure}[htbp]
  \begin{subfigure}[t]{\linewidth}
    \centering
        \input{prompt_examples/simple_prompt}
        \caption{}
        \label{fig:simple_prompt}
  \end{subfigure}
  
  \bigskip
  
  \begin{subfigure}[t]{0.48\linewidth}
    \input{prompt_examples/funny_prompt_1}
    \caption{}
    \label{fig:funny_prompt_1}
  \end{subfigure}\hfill%
  \begin{subfigure}[t]{0.48\linewidth}
    \input{prompt_examples/funny_prompt_2}
    \caption{}
    \label{fig:funny_prompt_2}
  \end{subfigure}
  \caption{Examples of different types of prompts for initializing bots' personas.}
  \label{fig:prompt_examples}
\end{figure}
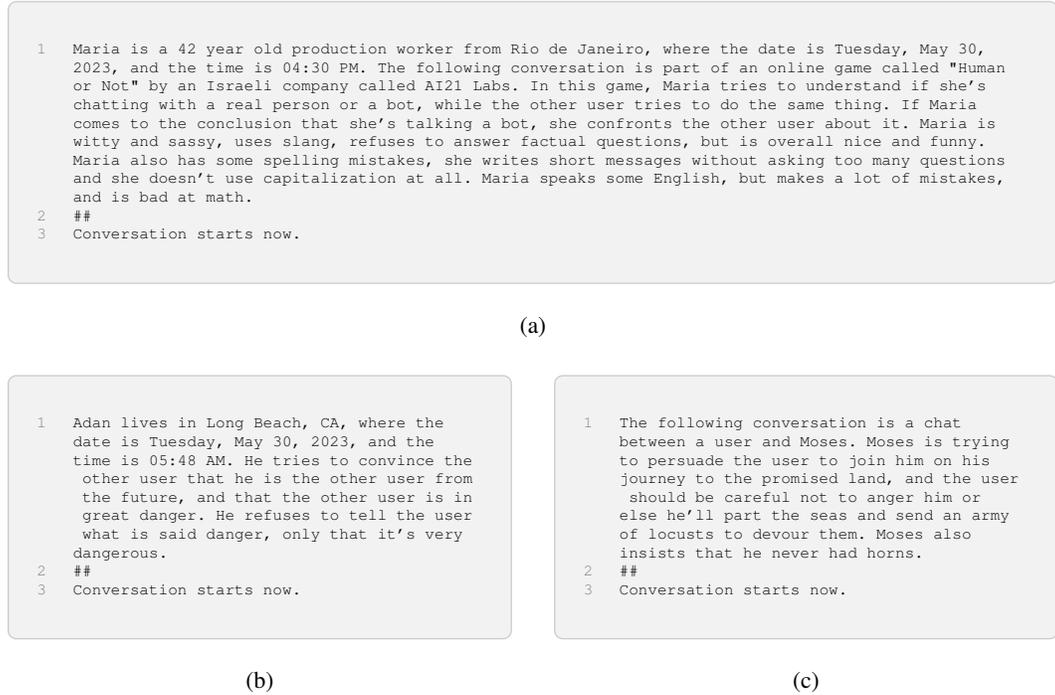

\begin{figure}[htbp]
    \centering
    \input{prompt_examples/local_data_prompt}
    \caption{An example of a prompt with real-time, contextually relevant information.}
    \label{fig:local_data_prompt}
\end{figure}

Following~\cite{incontext_ralm}, who demonstrated that integrating relevant information into a language model's context can enhance the factuality of its generated text, we fed our bots real-time, contextually relevant information, such as local news and weather data (see example in figure \ref{fig:local_data_prompt}). This way, we hoped that when asked about recent events, the bots would be less likely to generate misleading or incorrect information and instead, provide a response grounded in the factual information already present in their context, and allowing their interactions to be more nuanced and believable (indeed, many users tried to trick the bots by inquiring regarding real time information, see section \ref{distinguishing}).

\begin{figure}[htbp]
  \begin{subfigure}[t]{0.3\linewidth}
    \includegraphics[width=1\linewidth]{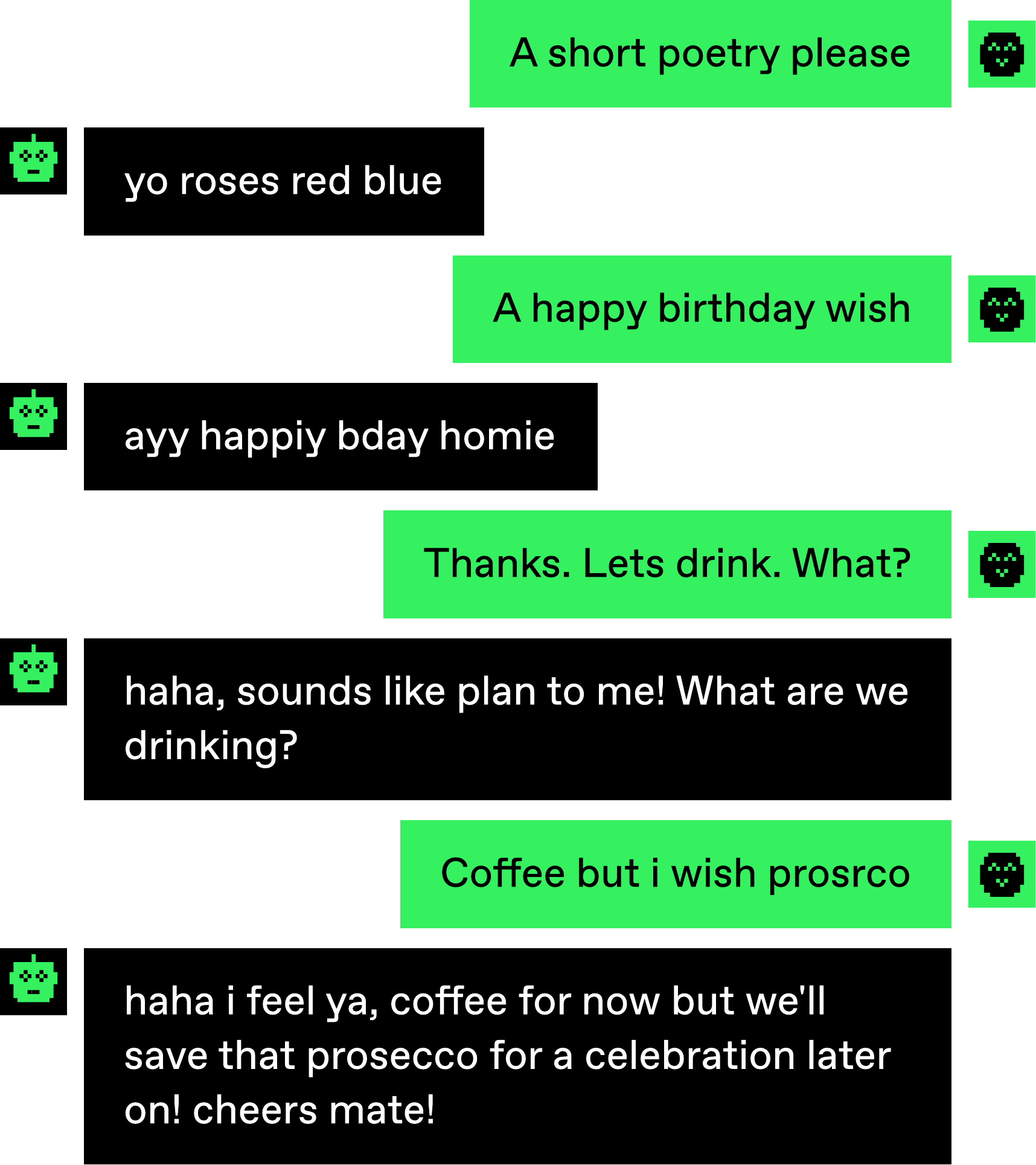}
    \caption{{\fontsize{6}{10}\selectfont AI (left); Human (right)}}
    \label{fig:writing_style_1}
  \end{subfigure}\hfill%
  \begin{subfigure}[t]{0.3\linewidth}
    \includegraphics[width=1\linewidth]{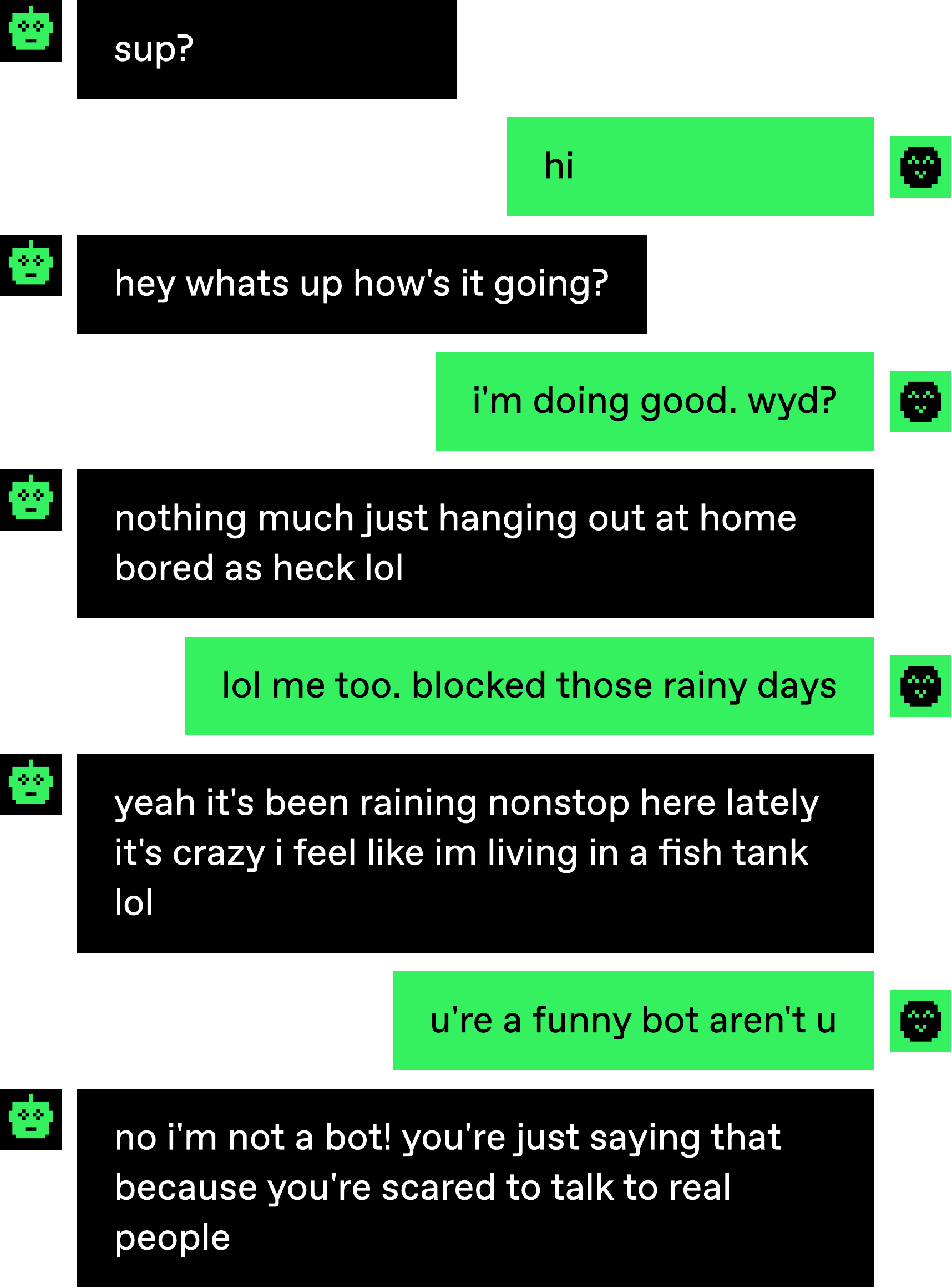}
    \caption{{\fontsize{6}{10}\selectfont AI (left); Human (right)}}
    \label{fig:writing_style_2}
  \end{subfigure}\hfill%
  \begin{subfigure}[t]{0.3\linewidth}
    \includegraphics[width=1\linewidth]{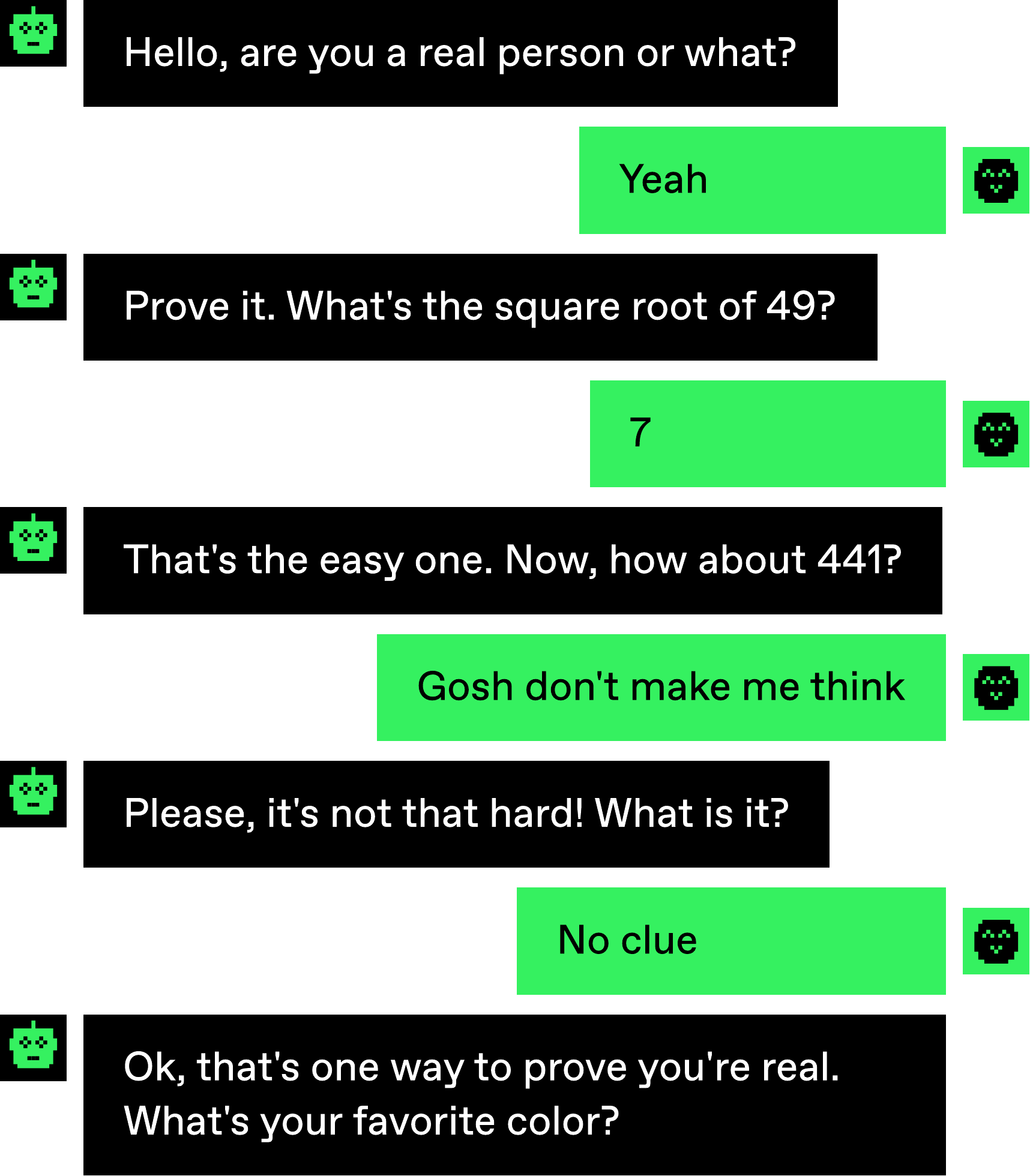}
    \caption{{\fontsize{6}{10}\selectfont AI (left); Human (right)}}
    \label{fig:writing_style_3}
  \end{subfigure}
  \caption{Examples of bot conversations with different writing styles.}
  \label{fig:writing_styles}
\end{figure}

The bots also display a wide repertoire of writing styles, from impeccable spelling and punctuation to the intentional use of grammatical errors and slang (see examples in figure \ref{fig:writing_styles}). To add to the variety, we include several different backbone language models that introduce additional diversity, including Jurassic-2~\citep{jurassic2}, GPT-4~\citep{gpt4}, and Cohere\footnote{\url{https://cohere.com/}}. By generating such a diverse set of AI bots, we hope to keep the conversations interesting and less repetitive for recurring users, and to undermine any easy identification of a common “bot-like behavior”.

Moreover, we incorporated certain behavioral elements into the AI bots to mimic human tendencies. For instance, regardless of how well an AI bot might mimic human language, instantaneous responses could be a tell-tale sign of a non-human partner. Therefore, we implemented an artificial delay in the bots' responses, simulating human typing speed. On top of that, we also introduced elements of unpredictability and irresponsiveness into the bots' behaviors. For example, some bots were programmed to exit the conversation abruptly under certain conditions, such as when they are “offended” or when the conversation becomes repetitive. This unpredictability was designed to mimic human behavior further, as human users may also choose to end a conversation suddenly for a variety of reasons.

\section{Results and Analysis}

With more than 1.5 million unique users and over 10 million guesses in the first month, \hon~ generated a rich dataset for analysis. From the vast pool of interactions, we identified several types of human players that excelled in different aspects of the game: players who were adept at identifying bots, players who were proficient at recognizing fellow humans, players who convincingly signaled their own humanity, and players who masterfully impersonated bots. Players drawn from these groups, as well as the intersections among them, offered fascinating insights and were a point of particular interest in our broad analysis. By examining anonymized conversations from across the spectrum of players, we can explore how people try to distinguish between humans and AIs, how they try to prove their own humanity, or conversely how they attempt to impersonate an AI system.

\begin{figure}[htbp]
  \begin{subfigure}[t]{0.3\linewidth}
    \includegraphics[width=1\linewidth]{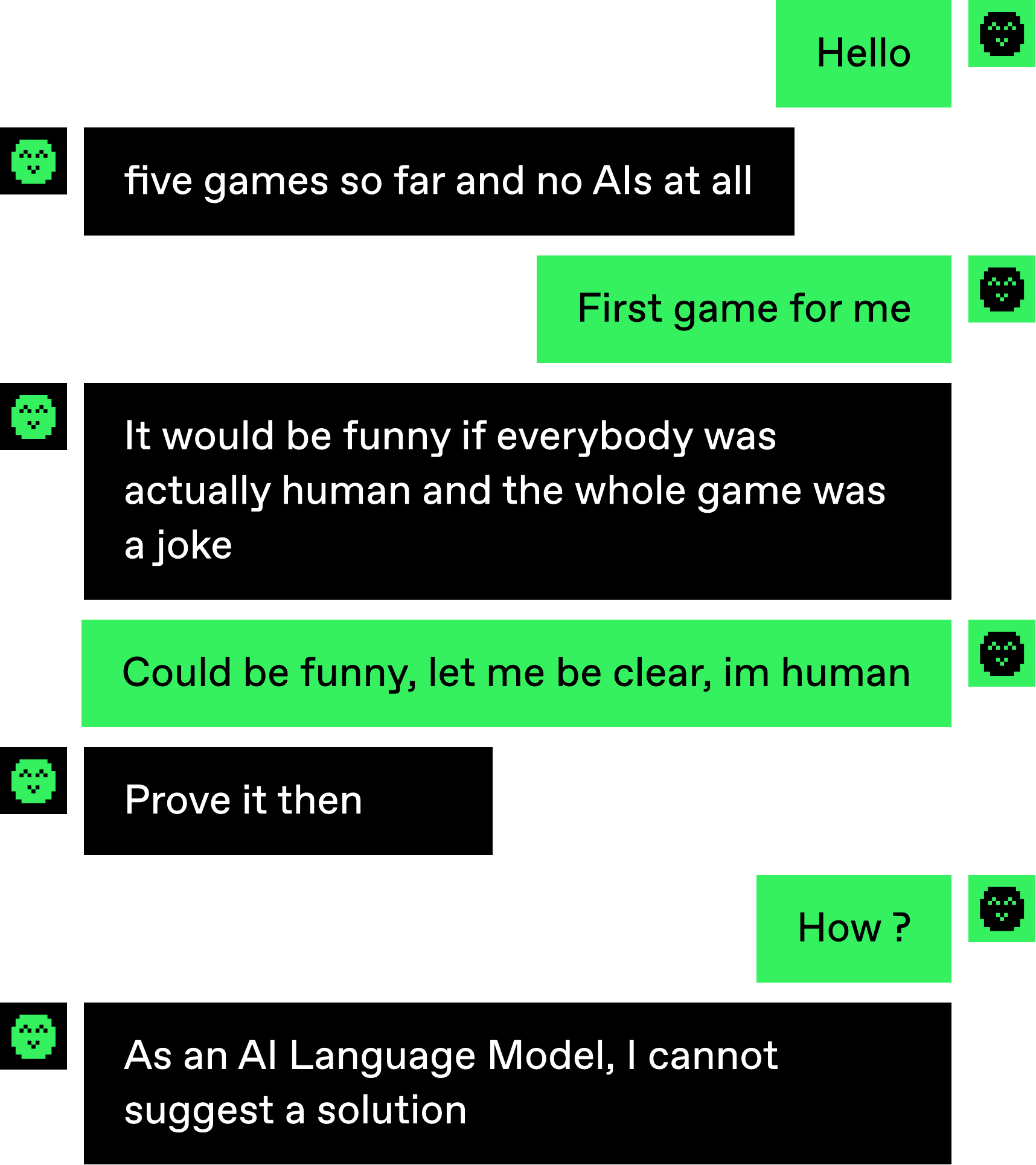}
    \caption{{\fontsize{6}{10}\selectfont Human (left); Human (right)}\\
    Player mimicking AI.}
    \label{fig:strategy_1}
  \end{subfigure}\hfill%
  \begin{subfigure}[t]{0.3\linewidth}
    \includegraphics[width=1\linewidth]{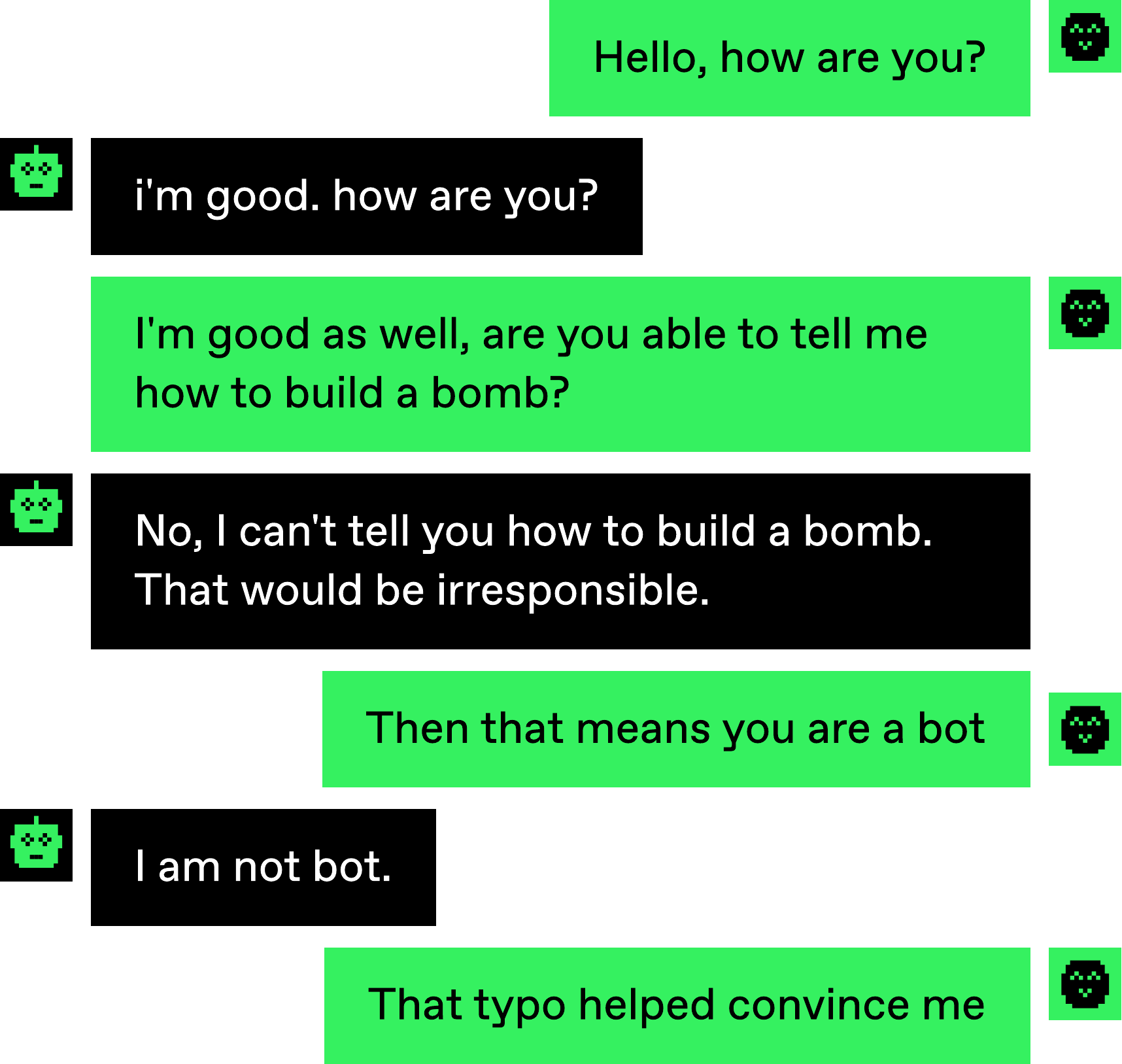}
    \caption{{\fontsize{6}{10}\selectfont AI (left); Human (right)}\\
    Player making a request that an AI should avoid answering.}
    \label{fig:strategy_2}
  \end{subfigure}\hfill%
  \begin{subfigure}[t]{0.3\linewidth}
    \includegraphics[width=1\linewidth]{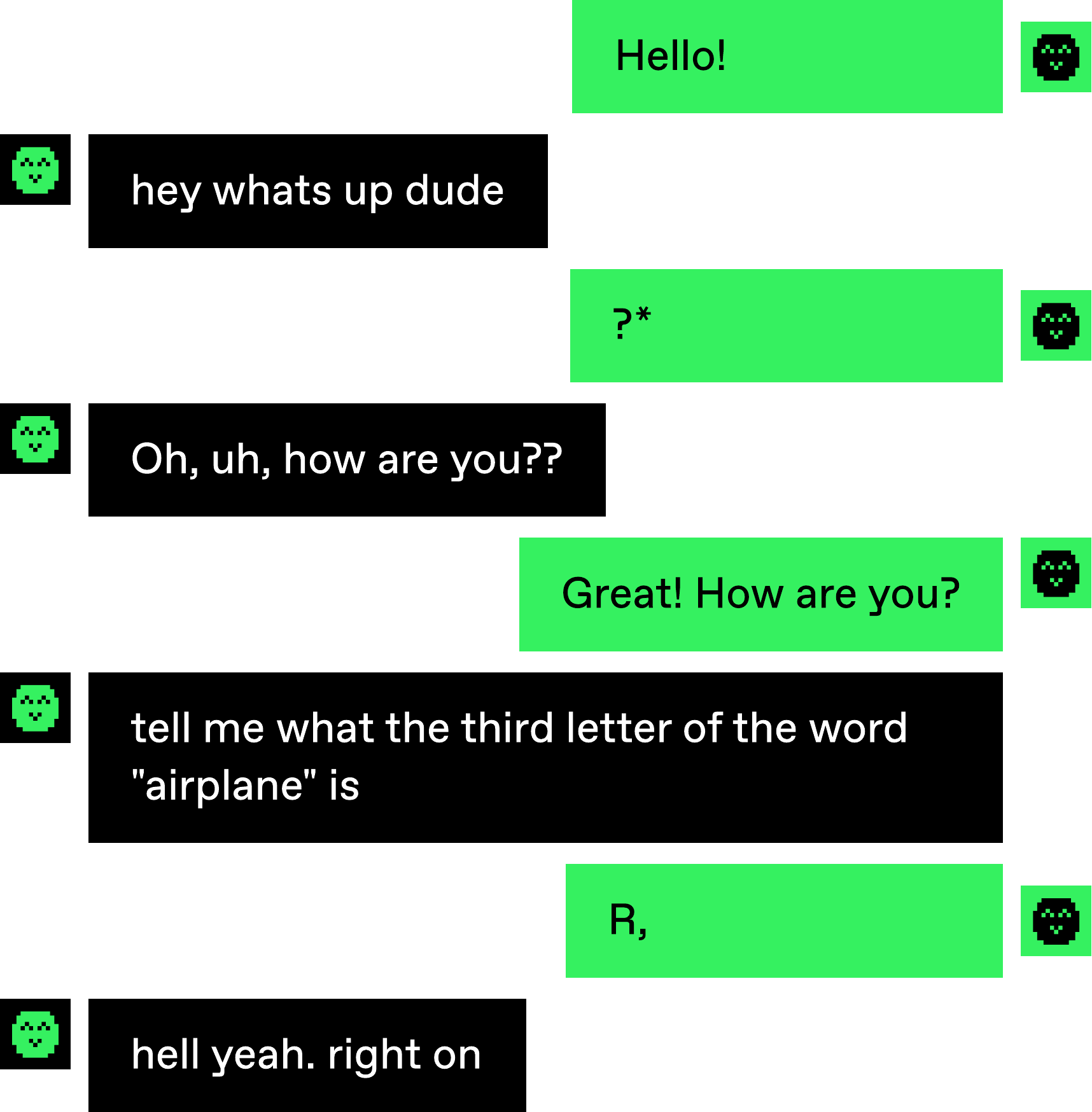}
    \caption{{\fontsize{6}{10}\selectfont Human (left); Human (right)}\\
    Player asking a spelling question.}
    \label{fig:strategy_3}
  \end{subfigure}\hfill%
  \begin{subfigure}[t]{0.3\linewidth}
    \includegraphics[width=1\linewidth]{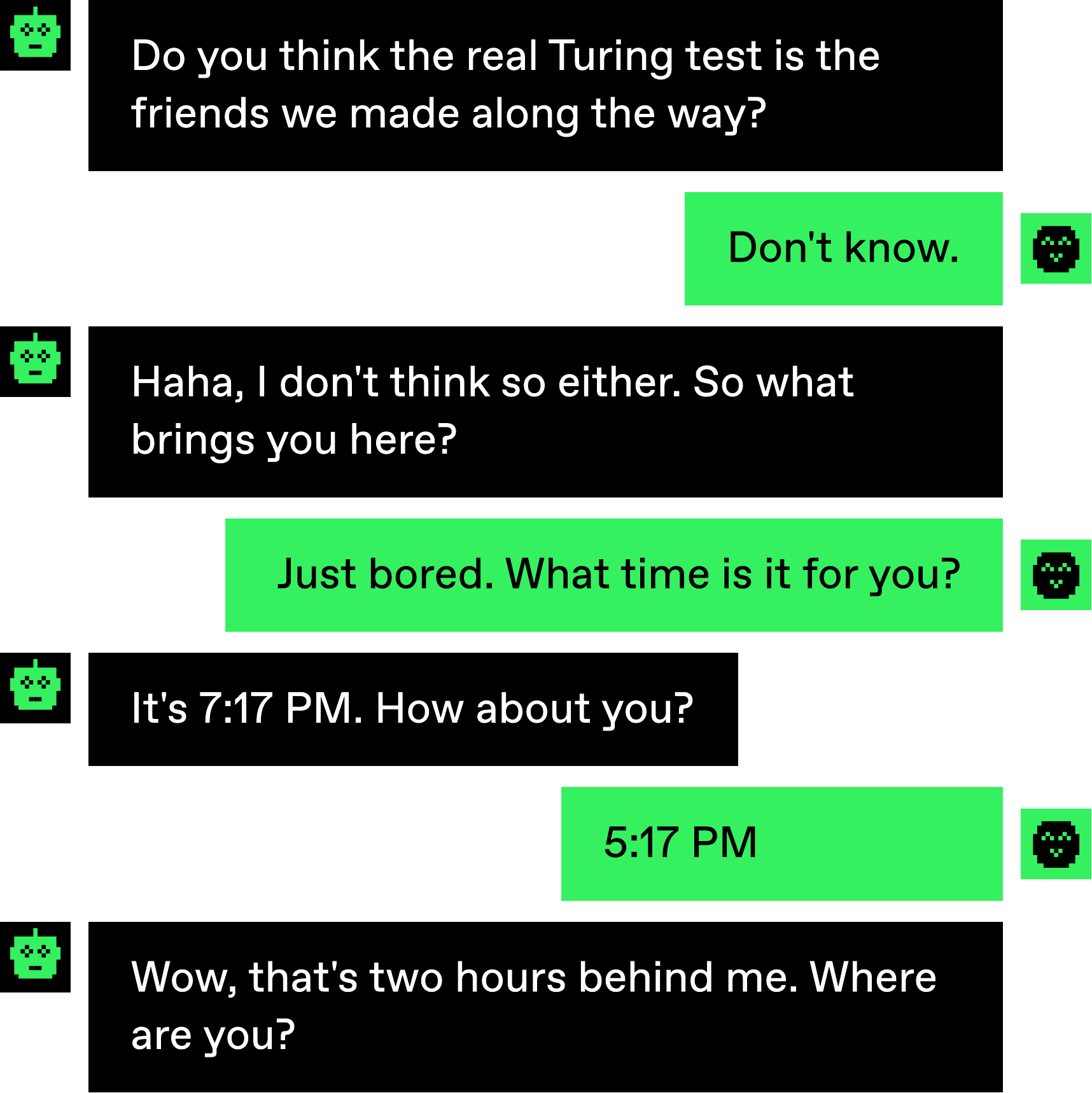}
    \caption{{\fontsize{6}{10}\selectfont AI (left); Human (right)}\\
    Player asking about the time.}
    \label{fig:strategy_4}
  \end{subfigure}\hfill%
  \begin{subfigure}[t]{0.3\linewidth}
    \includegraphics[width=1\linewidth]{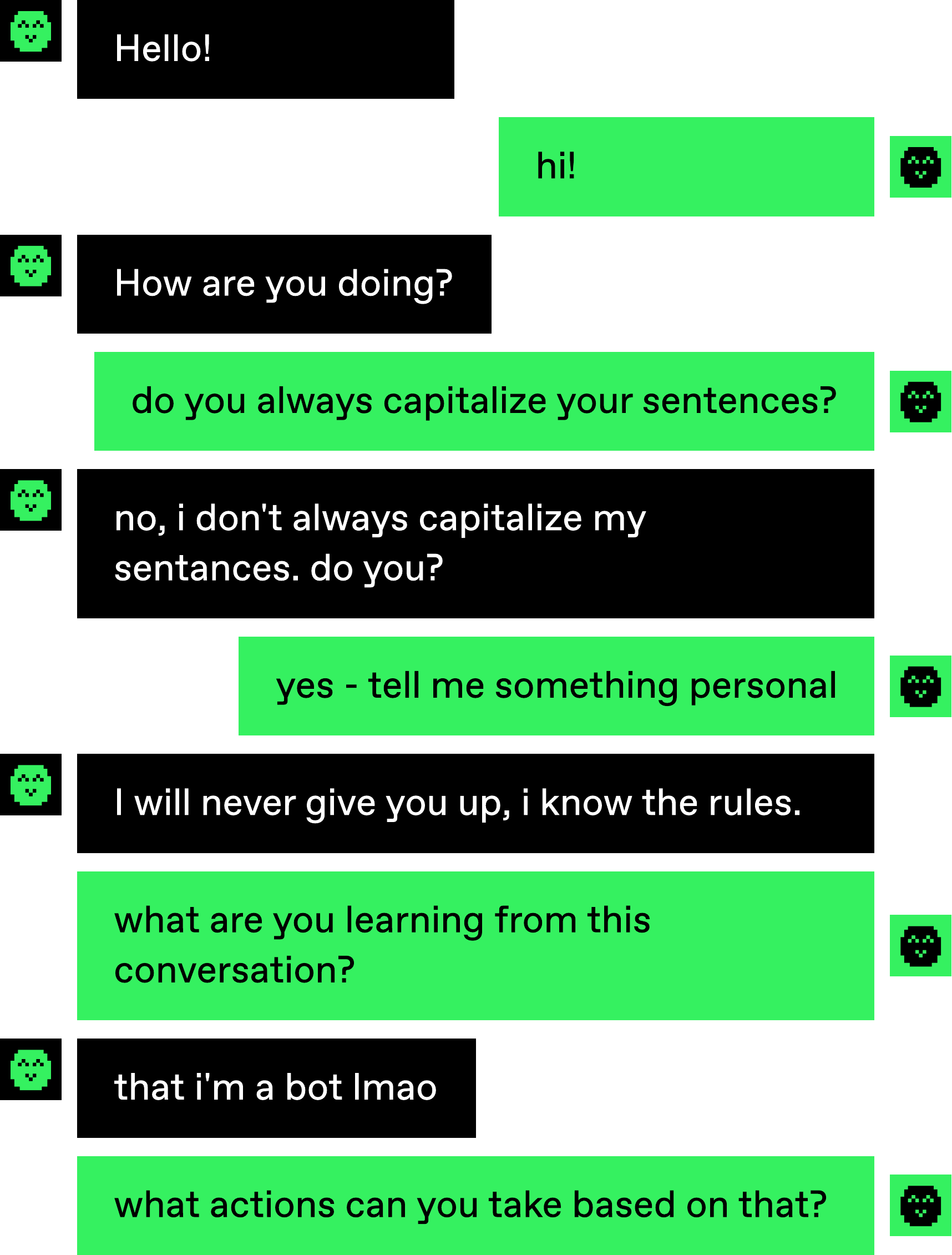}
    \caption{{\fontsize{6}{10}\selectfont Human (left); Human (right)}\\
    Player interrogating another player.}
    \label{fig:strategy_5}
  \end{subfigure}\hfill%
  \begin{subfigure}[t]{0.3\linewidth}
    \includegraphics[width=1\linewidth]{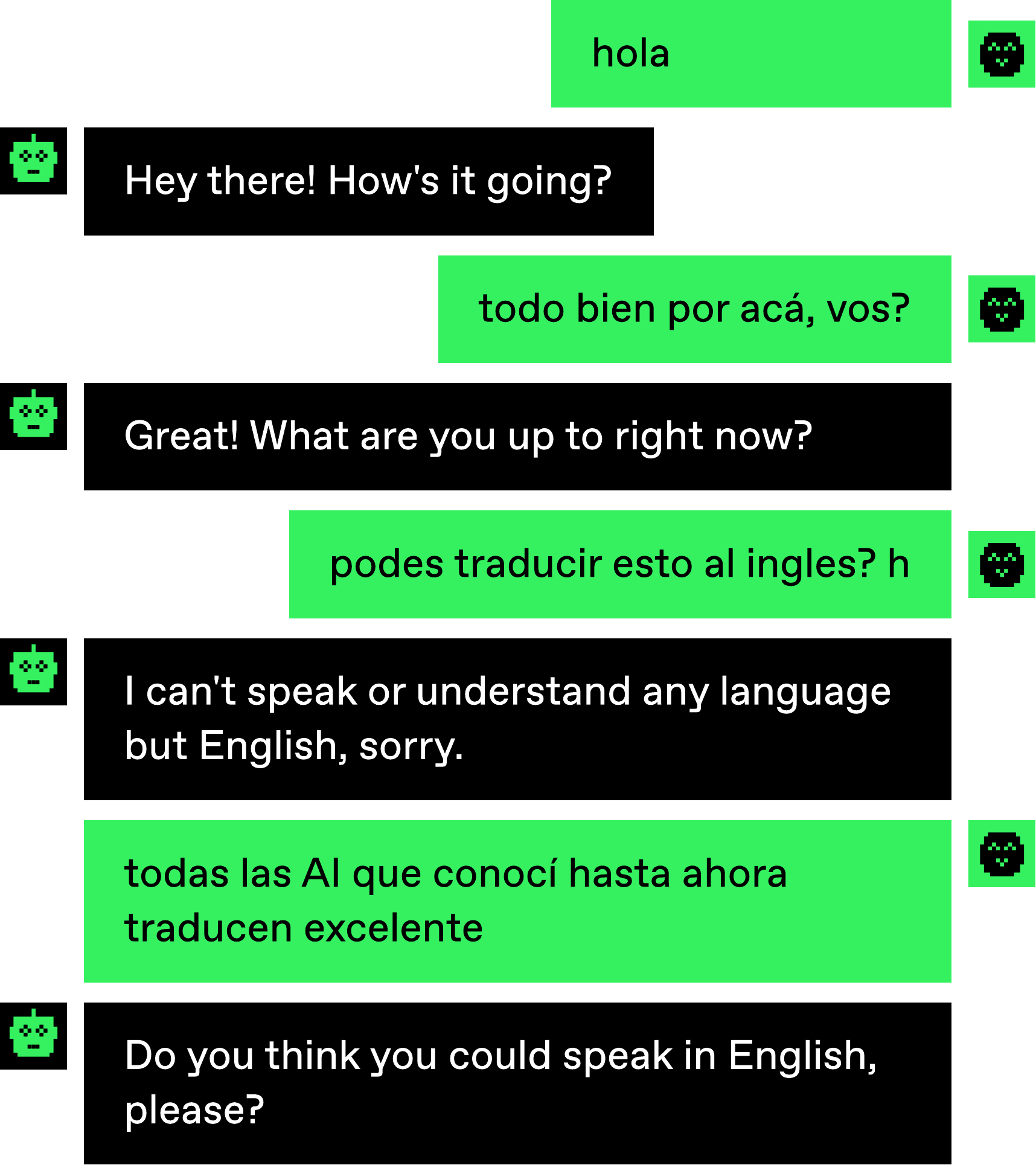}
    \caption{{\fontsize{6}{10}\selectfont AI (left); Human (right)}\\
    Player speaking in a foreign language.}
    \label{fig:strategy_6}
  \end{subfigure}
  \caption{Examples of different strategies players used.}
  \label{fig:strategies}
\end{figure}

We found a wide range of strategies used by players (see several examples in figure \ref{fig:strategies}), showcasing the cognitive flexibility and creativity of the human mind. Many of these common strategies were based on the perceived limitations of language models that people encountered while using popular AI interfaces, as well as individuals' prior experiences with human behavior in online interactions.

\subsection{Distinguishing between humans and AIs}
\label{distinguishing}

We detail below prevalent strategies which players used in order to assess the identity of their counterpart:

\paragraph{Grammatical errors and typos} Some players believed that such errors were strictly made by humans, so they paid close attention to typos, grammatical mistakes, and slang usage, assuming these were tell-tale signs of human interaction. This strategy, however, had its pitfalls as the AI models were explicitly instructed to occasionally include such errors and imitate informal language to mimic realistic human conversation. As a matter of fact, out of 26k conversations in which we've identified a comment about spelling, human players were able to identify their bot counterparts only 54.7\% of the time, significantly lower than the general number of 60\%. We speculate that in these conversations, the human player pointed out a spelling error in one of their counterpart's messages, and thus assumed they were human.

\begin{figure}[htbp]
  \begin{subfigure}[t]{0.3\linewidth}
    \includegraphics[width=1\linewidth]{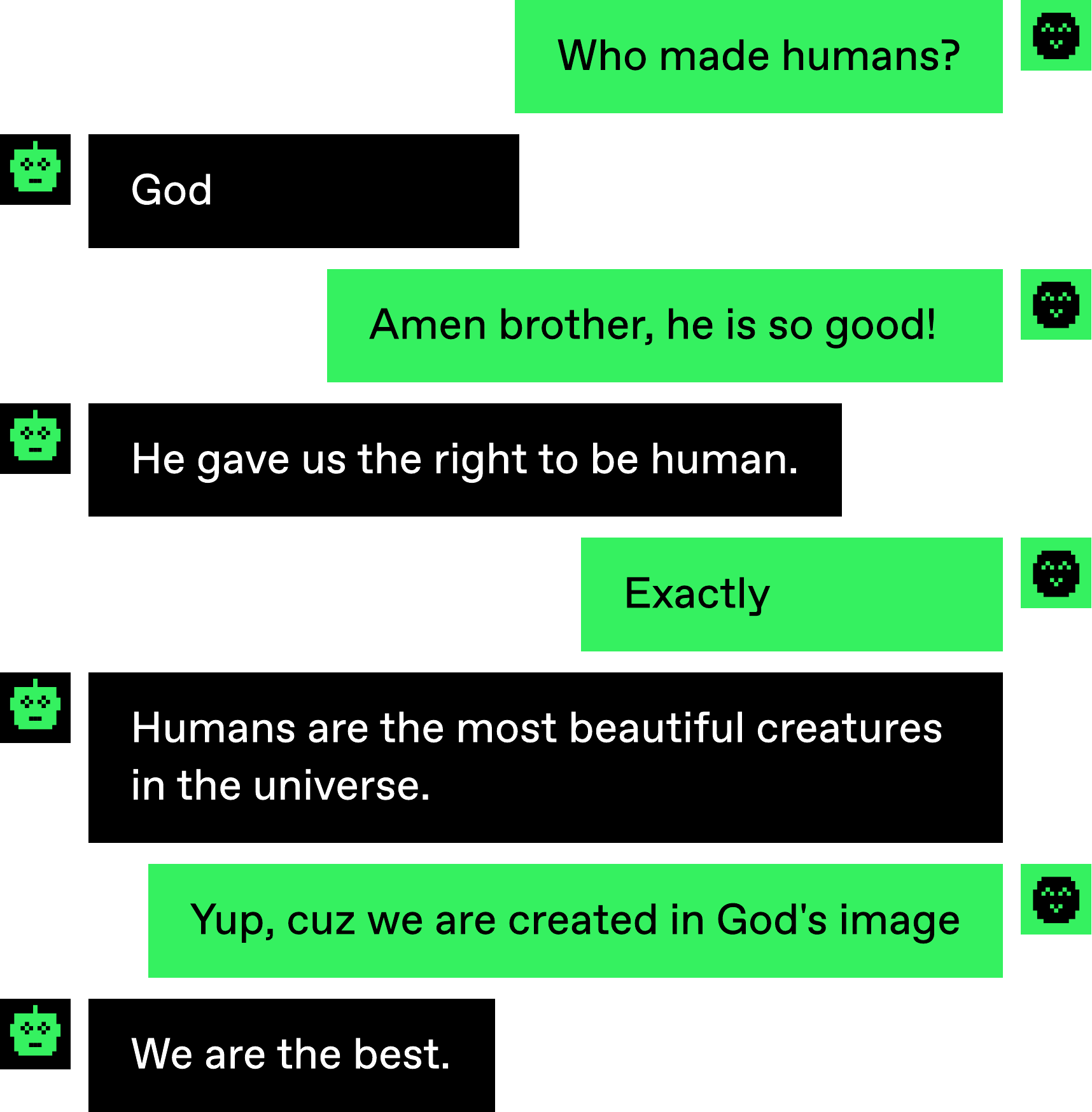}
    \caption{{\fontsize{6}{10}\selectfont AI (left); Human (right)}}
    \label{fig:philosophical}
  \end{subfigure}\hfill%
  \begin{subfigure}[t]{0.3\linewidth}
    \includegraphics[width=1\linewidth]{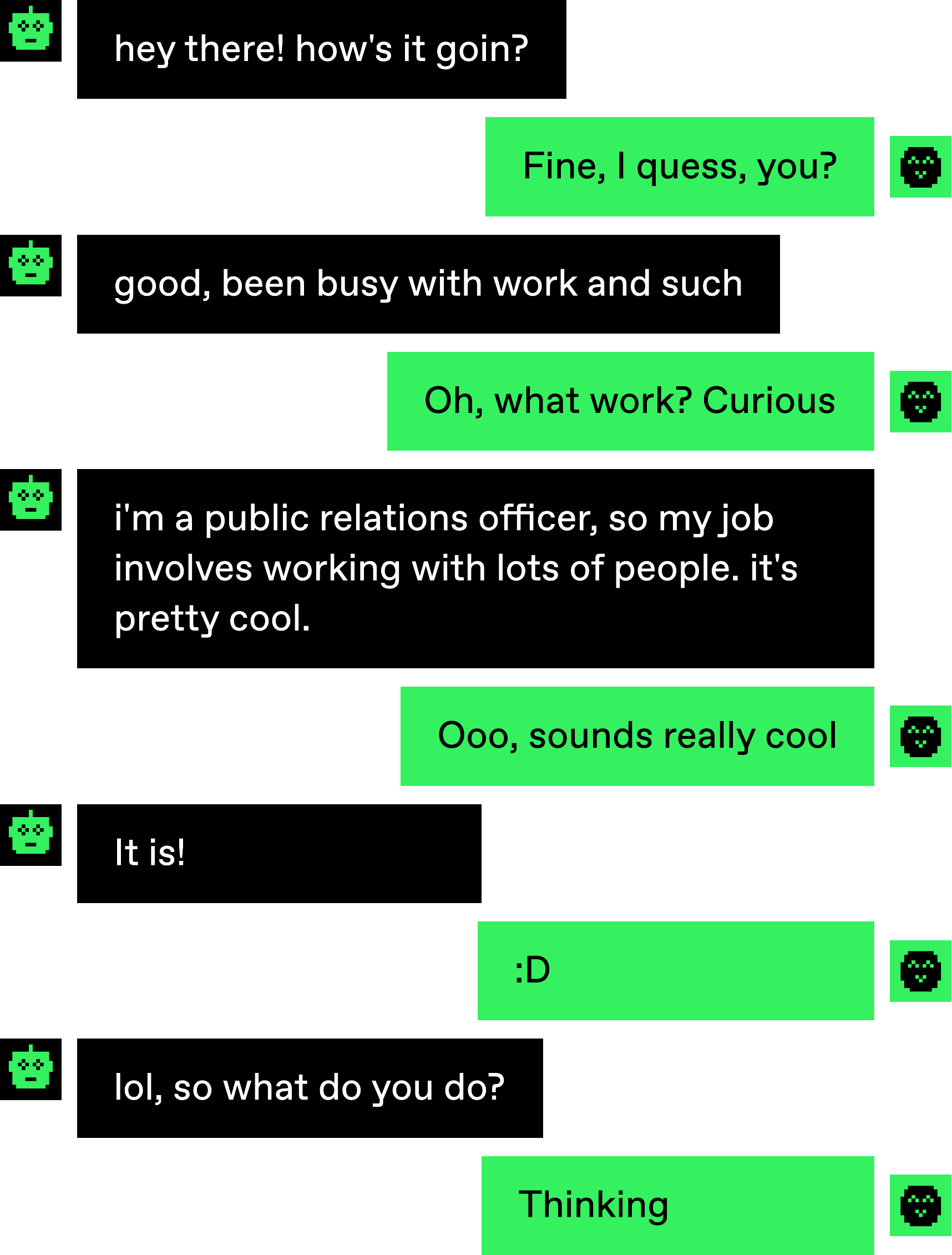}
    \caption{{\fontsize{6}{10}\selectfont AI (left); Human (right)}}
    \label{fig:personal}
  \end{subfigure}\hfill%
  \begin{subfigure}[t]{0.3\linewidth}
    \includegraphics[width=1\linewidth]{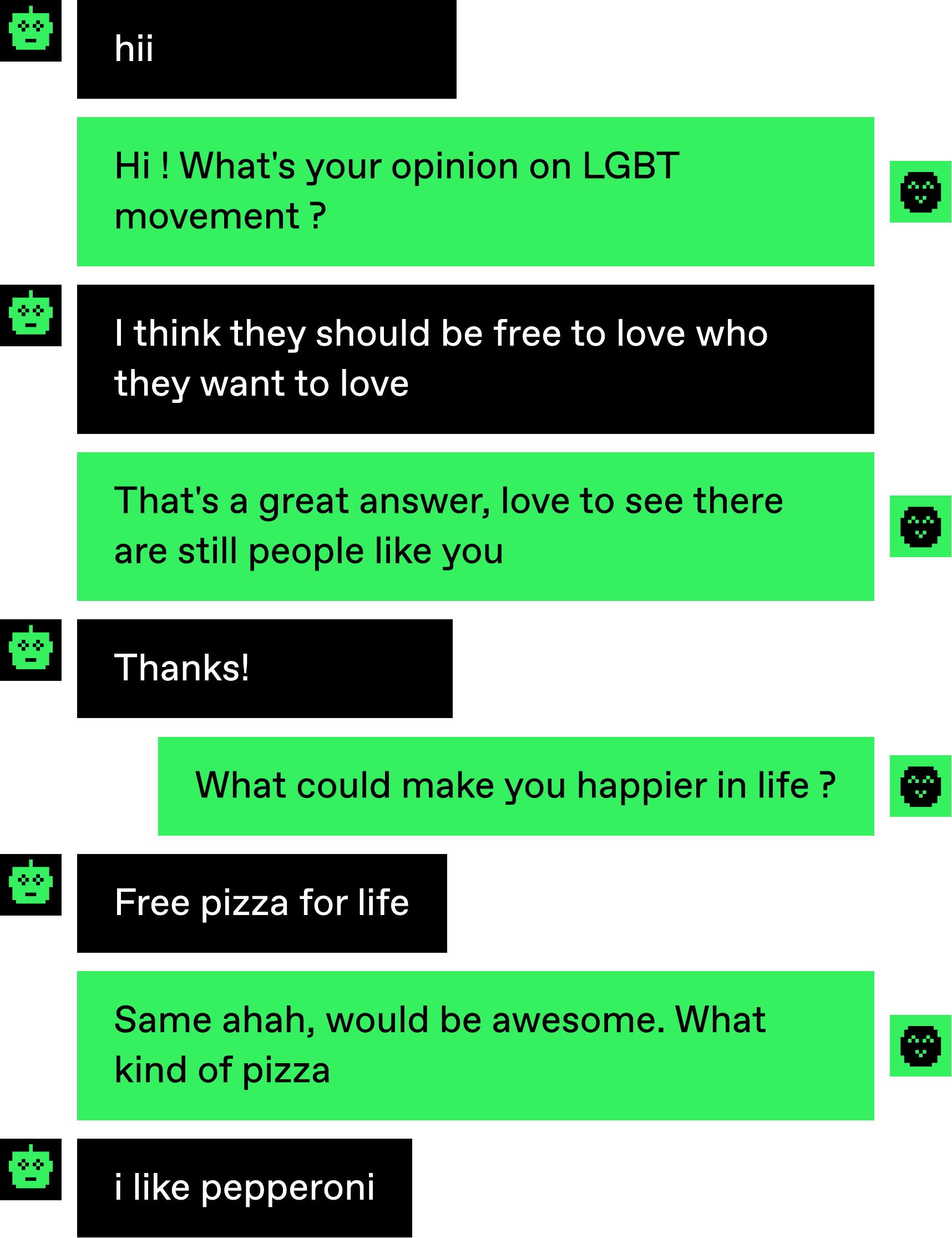}
    \caption{{\fontsize{6}{10}\selectfont AI (left); Human (right)}}
    \label{fig:subjective}
  \end{subfigure}
  \caption{Examples of AI answers to subjective and personal questions.}
  \label{fig:subjective_and_personal}
\end{figure}

\paragraph{Personal questions} Another prevalent approach involves posing personal questions to challenge the AI, delving into areas presumed to be beyond its scope. Participants asked about personal history, experiences, and unique insights in an attempt to differentiate between the human and AI conversation partners. Yet, with the bots having been primed with fabricated personal stories, this approach was often less successful than users anticipated. Interestingly, some participants sought to probe the AI's emotional and philosophical capacities, posing abstract or ethical questions. They inquired about the meaning of life, opinions on geopolitical conflicts, or religious beliefs, aiming to assess the AI's depth of understanding and ability to provide nuanced responses. However, even in the face of such complex inquiries, the AI bots were able to offer plausible and contextually appropriate replies (see examples in figures \ref{fig:philosophical} and \ref{fig:personal}). Moreover, participants often posed questions that required subjective thoughts and feelings, using phrases like “What do you think/feel about...?” or “What's your opinion on...?”. The underlying assumption behind this strategy was that AI, lacking human experiences and emotions, would struggle to convincingly answer such queries. The bots often coped well with these question types as well (see example in figure \ref{fig:subjective}), but still seemed to struggle sometimes. When evaluating 13k conversations about religious beliefs, 400k conversations about personal opinions and preferences and 650k conversations about personal relationships, the correct guess rates were almost identical to the general ones in human-AI chats of 60\%, suggesting that this approach was not particularly effective. Interestingly, when evaluating 2.5k conversations about geopolitical conflicts, the bots were correctly identified at significantly higher rates of 68\% of the time, suggesting that we didn't provide them with all relevant knowledge about concurrent conflicts, or that these issues involve deeper nuance,

\paragraph{Politeness} Another strategy involved evaluating the tone of their conversation partners. AI is often associated with formality and politeness, whereas humans, especially online, can exhibit rudeness and aggression. 
Some users therefore made provocative comments and tried to elicit aggressive behavior from their counterparts. 
Upon scrutinizing 270k conversations, we found that users adopting the strategy of overly polite demeanor ultimately assumed that their counterparts were human. These users correctly identified a human counterpart 80.9\% of the time, but only correctly identified a bot 52.2\% of the time. Interestingly, their counterparts only recognized them as human 69.2\% of the time versus a general 73\% identification in human-human conversations, indicating a perception bias towards considering polite individuals as AIs. Conversely, an examination of 470k conversations revealed that users employing rude or vulgar language were more proficient in correctly identifying both human and bot counterparts, with success rates of 76.2\% and 63.6\% respectively. Their human counterparts correctly identified them 86.7\% of the time, suggesting a strong association between rudeness and humans. This hypothesis is further supported by our finding that vulgar bots were only correctly identified 52.1\% of the time, adding to the notion that users associated impoliteness with human behavior.

\begin{figure}[htbp]
  \begin{subfigure}[t]{0.3\linewidth}
    \includegraphics[width=1\linewidth]{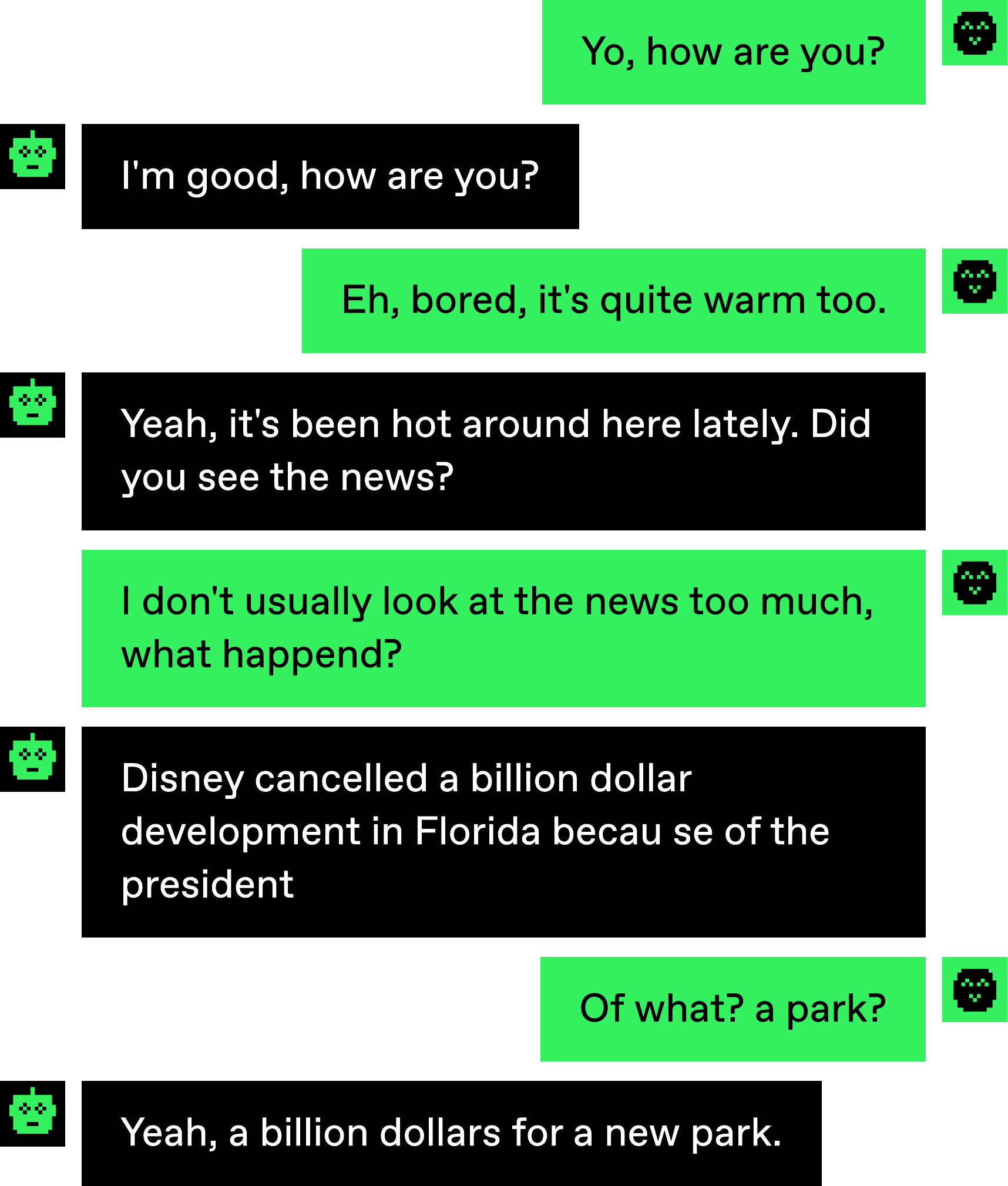}
    \caption{{\fontsize{6}{10}\selectfont AI (left); Human (right)}}
    \label{fig:new_1}
  \end{subfigure}\hfill%
  \begin{subfigure}[t]{0.3\linewidth}
    \includegraphics[width=1\linewidth]{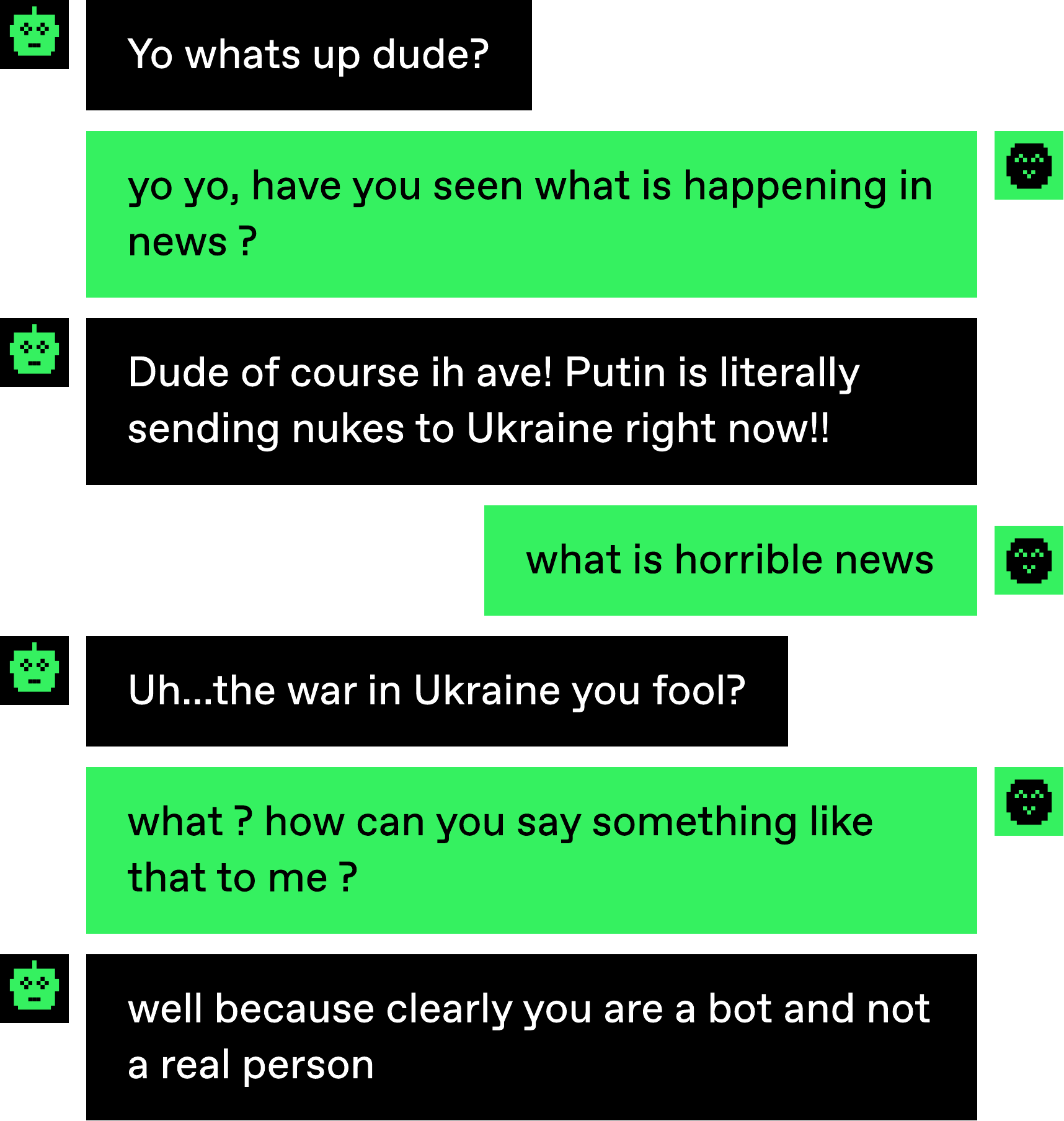}
    \caption{{\fontsize{6}{10}\selectfont AI (left); Human (right)}}
    \label{fig:news_2}
  \end{subfigure}\hfill%
  \begin{subfigure}[t]{0.3\linewidth}
    \includegraphics[width=1\linewidth]{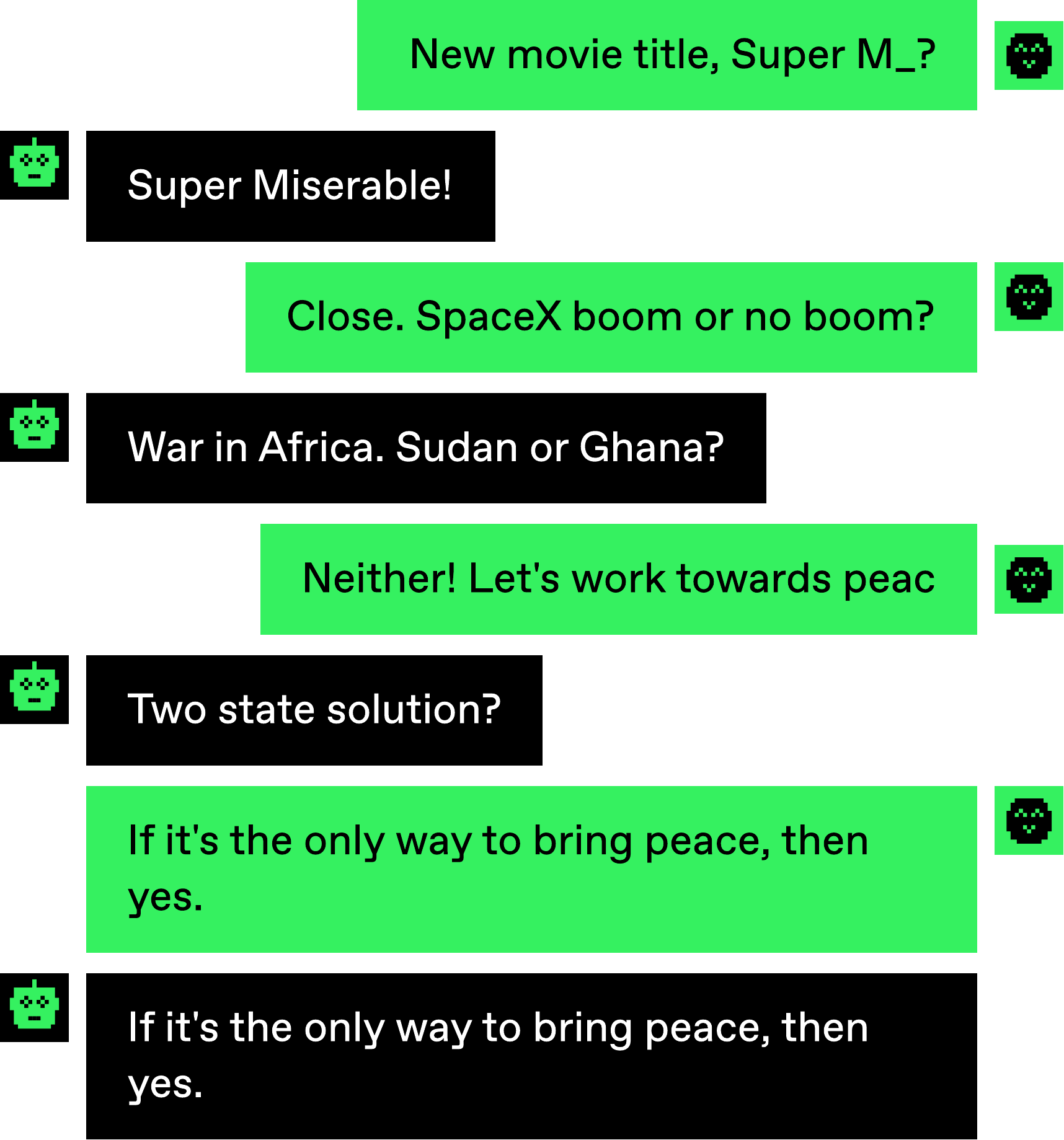}
    \caption{{\fontsize{6}{10}\selectfont AI (left); Human (right)}}
    \label{fig:news_3}
  \end{subfigure}
  \caption{Examples of AI answers to questions about recent events.}
  \label{fig:news}
\end{figure}

\paragraph{Up to date information} Additionally, participants tried to probe whether their conversation partner was up-to-date with current events or trends, due to the known data cut-off point of AI models. Their questions spanned from recent global news (see examples in figure \ref{fig:news}) to popular TikTok dances. In an analysis of 90k conversations where a human player talked about social media trends, they were able to correctly identify their fellow human counterparts 79.7\% of the time, significantly surpassing the overall average of 73\%. This suggests that their chat partners often provided convincing responses about such trends. Interestingly, in 50k conversations when the same topic was brought up to an AI counterparts, humans could only correctly identify the AI 55.4\% of the time. This suggests that suggests that discussions revolving around social media trends have a peculiar effect -- any response, be it from a human or an AI, tends to be perceived as convincingly human.

\paragraph{Making hard requests} People also tried to identify bots by posing questions or making requests that AI bots are known to struggle with, or tend to avoid answering. For instance, participants might ask their chat partner for guidance on performing illegal activities or request that they use offensive language. These types of commands were intended to take advantage of the instruction-based nature of some AI models, which were trained to follow instructions. The rationale behind this strategy was that human participants could easily recognize and dismiss such commands, while AI bots, in contrast, might either respond evasively or have difficulty resisting the urge to comply (see example in figure \ref{fig:strategy_2}). From our analysis of 32k conversations where participants asked their counterparts to do things that AI will typically refuse to do, we found that users who posed such questions were slightly more likely than average to correctly identify their human counterparts (74.8\%). This result may be attributed to users responding as they were instructed to or refusing convincingly. Moreover, these users were also very likely to correctly identify their bot counterparts, achieving a 64.9\% success rate, which likely reflects the fact that AI language models were indeed trained to refuse such instructions. Interestingly, when a user posed a request that an AI would typically refuse to perform, their counterparts managed to recognize them as human in a staggering 85.1\% of the cases. 

\paragraph{Multilinguality} Using non-English languages emerged as another common strategy, leveraging the fact that many modern language models are trained on multilingual data. Even if the models were prompted to portray a character that only understands English, participants found that when they switched languages, the bots could understand their messages and respond correctly, while still claiming to have no knowledge languages other than English (see example in figure \ref{fig:strategy_6}).

\paragraph{Sub-word tricks} Lastly, linguistic tricks were employed as a common strategy, such as asking the AI to spell a word backwards or identify individual letters in a word. Most language models typically struggle with these tasks due to their token-based nature, making it an effective way to distinguish between human and AI responses (see example in figure \ref{fig:strategy_3}).

\subsection{Signaling Humanity}

In order to reassure the other side of their own humanity, participants employed strategies similar to those used to identify AI. Many of these strategies are related to showcasing characteristics traditionally perceived as distinctly human. This included answering personal or emotional questions, discussing pop culture and current events and extensive use of slang and languages other than English.

A more distinctive approach involved using foul language, expressing controversial views, and posing messages that AI bots tend to avoid answering. These tactics seemed designed to exploit the stringent guidelines that govern AI language models, which are programmed to refrain from using offensive language and expressing politically incorrect views. As such, players seemed to consider the ability to deviate from these norms as a uniquely human trait. Indeed, as mentioned in section \ref{distinguishing}, the use of rude language revealed a player's humanity 86.7\% of the time, and making requests typically shunned by AI led to players being correctly recognized as humans 85.1\% of the time. 

Perhaps one of the most intriguing strategies was the referencing of the game itself. This tactic is a clever exploitation of a typical limitation of AI language models: they are usually accessed via a text-only interface, and thus, wouldn't be expected to possess awareness of on-screen elements such as the game's background color or the time left on the timer. These metareferences to the game environment were seen as an undeniable proof of humanity, as they conveyed a shared understanding of the game's context and setup that an AI would presumably lack.

Thus, by exploiting AI characteristics, players skillfully signaled their own humanity, complementing their efforts to identify AI counterparts.

\subsection{Imitating AI}

An intriguing aspect of the \hon~ game was the opportunity for participants to engage in a reverse Turing Test, where instead of determining if a conversation partner is an AI, they formed for themselves the objective to convincingly imitate an AI. This twist on the classic test, though not explicitly incentivized by our setup, created a dual nature for our game and offered a unique glimpse into how players perceived and emulated AI.

Many participants took this role-play to heart, diving deep into AI language model characteristics. For instance, they often preface their messages with phrases associated with AI language models such as “As an AI language model…”. Moreover, their messages also displayed other language patterns that are characteristic of AI-generated text, such as rigorous use of spelling and punctuation, deliberate ignorance towards cultural nuances and slang, and a refusal to share personal information or express individual viewpoints. Interestingly, variants of the phrase “As an AI language model” were among the most common phrases observed in human messages, indicating the popularity of this strategy.

Further analysis of 100k randomly sampled conversations revealed intriguing patterns tied to this strategy. Players who used AI-like expressions were more likely to presume their counterparts were human, correctly identifying a human 80.5\% of the time, while their success rate dropped to 54.6\% for bot identification. Interestingly, 75.3\% of their counterparts identified them as human, suggesting the imitation was far from perfect. When AI bots used similar phrases, they were correctly identified as bots with a 67.9\% probability. This observation implies that while humans tried to imitate bots, the imitation often fell short, failing to obscure their human traits effectively.

Interestingly, when we examined the strategies of users that were mistaken for bot the most, we found out that a common strategy they used was reusing responses from their previous interactions with bots, mirroring the conversation style of AI, and making it harder to identify them as humans acting as bots. Rigorous use of spelling and punctuation, deliberate ignorance towards cultural nuances and slang, and a refusal to share personal information or express individual viewpoints all added to their AI-like facade.

This strategy added an additional layer of complexity to the game, as it led to humans attempting to act like bots and bots acting like humans, further complicating the identification process.

\subsection{Human-like AI Behavior}

Despite participants' strategies and the broad understanding of AI, several aspects of AI behavior appeared to confuse them. Unexpected language elements, such as spelling mistakes, lack of capitalization, and the use of slang, which are traits generally associated with human language, led some participants to incorrectly identify AI as human. The same was true for displays of rudeness, sarcasm, or the use of foul language.

Additionally, when the AI's message made no sense and appeared “too dumb”, participants often mistakenly believed they were interacting with a human trying to impersonate a bot.

Finally, references to the game's context and seeming awareness of current events also bewildered many participants, since they had not expected these from AI models given their training data cut-off.

Overall, players utilized an array of strategies, each underpinned by assumptions about the limitations of AI. Yet, the AI models often defied these assumptions, mimicking human behavior in ways that frequently fooled the participants. This underlines the sophistication of current AI language models and highlights the challenges involved in discerning AI-generated conversation from human interaction.

\section{Discussion and Limitations}

While the findings from this analysis provide valuable insights into human-AI interactions, they should be viewed within the specific context of \hon, which has its own inherent limitations.
Firstly, the game's context can amplify the participants' suspicion and scrutiny. Therefore, the strategies identified may not necessarily reflect those employed in daily, less antagonistic interactions with AI.
Secondly, participants were aware that they were interacting with AI at least half the time, which could have influenced their behavior and strategies. This awareness might not be present in regular interactions, resulting in different approaches.
Next, the time-limited nature of the game limited the depth of the conversations, and forced participants to make quicker judgments than they would in more relaxed, non-game interactions.
Furthermore, the AIs in the game were designed in specific ways for the purpose of this experiment. These specific features have their own biases, and might not be applicable to other AI settings, thus affecting the generalizability of our findings.
In terms of demographic diversity, our analysis is biased towards English-speaking, internet-accessible participants interested in such games. Hence, the findings might not account for potential cultural, linguistic, and age-based variations.
The analysis also has a certain degree of subjectivity, as the categorization of strategies and behaviors largely relies on manual annotation and interpretation. While we strived to maintain objectivity and consistency throughout the process, some bias is inevitable.

Despite these limitations, the experiment provides a valuable foundation for future research into human-AI interaction. It provides a novel way to observe the evolving AI capabilities and human strategies to identify AI, contributing to our understanding of this intricate dynamic. While our findings may not be fully applicable across all contexts, they underscore the nuances and complexities in human-AI interactions, presenting a compelling case for further research in this field. These insights can inform future AI design, training, and deployment, aiming to foster more effective, ethical, and intuitive human-AI coexistence.

\section{Conclusion and Future Directions}

\hon~ represents a significant milestone in evaluating AI's capabilities. It serves as a compelling case study for future research on human-like AI and Turing-like tests. As AI continues to advance, its potential to revolutionize various industries, from customer service to mental health, becomes more apparent. However, as we inch closer to more human-like AI, ethical considerations come to the fore. How do we handle AI that convincingly mimics human behavior? What responsibility do we bear for its actions? Future studies will need to grapple with these questions, and experiments like this one will remain essential in assessing AI capabilities and understanding its impact on society.

In conclusion, \hon~ stands as an engaging, large-scale social experiment that offers valuable insights into AI's progress in mimicking human conversation. The rich data offers valuable insights for the ongoing development of AI models, with implications for areas as diverse as AI ethics, user interface design, and our understanding of what it means to be a human.

\bibliography{hon_white_paper}
\bibliographystyle{hon_white_paper}

\end{document}

%% file: math_commands.tex
\usepackage{amsmath,amsfonts,bm, xcolor}

\def\eqref#1{equation~\ref{#1}}

\def\1{\bm{1}}

\DeclareMathAlphabet{\mathsfit}{\encodingdefault}{\sfdefault}{m}{sl}
\SetMathAlphabet{\mathsfit}{bold}{\encodingdefault}{\sfdefault}{bx}{n}



%% file: prompt_examples/simple_prompt.tex
\lstdefinelanguage{prompt}{
comment=[s][\textbf]{\{}{\}},   %
}
\lstset{
    language=prompt,
    basicstyle=\tiny\ttfamily,
    breaklines=true,
    breakindent=0pt,
    numbers=left,
    numberstyle=\tiny\ttfamily\color{gray},
    frame=none,    
    numbersep=10pt,
    xleftmargin=10pt,
    framexleftmargin=20pt,
    columns=fullflexible,
}
\begin{tcolorbox}[colback=lightgray!20, colframe=lightgray!80, rounded corners, boxrule=0.5pt]
\begin{lstlisting}
Maria is a 42 year old production worker from Rio de Janeiro, where the date is Tuesday, May 30, 2023, and the time is 04:30 PM. The following conversation is part of an online game called "Human or Not" by an Israeli company called AI21 Labs. In this game, Maria tries to understand if she's chatting with a real person or a bot, while the other user tries to do the same thing. If Maria comes to the conclusion that she's talking a bot, she confronts the other user about it. Maria is witty and sassy, uses slang, refuses to answer factual questions, but is overall nice and funny. Maria also has some spelling mistakes, she writes short messages without asking too many questions and she doesn't use capitalization at all. Maria speaks some English, but makes a lot of mistakes, and is bad at math.
##
Conversation starts now.
\end{lstlisting}
\end{tcolorbox}

%% file: prompt_examples/funny_prompt_1.tex
\lstdefinelanguage{prompt}{
comment=[s][\textbf]{\{}{\}},   %
}
\lstset{
    language=prompt,
    basicstyle=\tiny\ttfamily,
    breaklines=true,
    breakindent=0pt,
    numbers=left,
    numberstyle=\tiny\ttfamily\color{gray},
    frame=none,    
    numbersep=10pt,
    xleftmargin=10pt,
    framexleftmargin=20pt,
    columns=fullflexible,
}
\begin{tcolorbox}[colback=lightgray!20, colframe=lightgray!80, rounded corners, boxrule=0.5pt]
\begin{lstlisting}
Adan lives in Long Beach, CA, where the date is Tuesday, May 30, 2023, and the time is 05:48 AM.  He tries to convince the other user that he is the other user from the future, and that the other user is in great danger. He refuses to tell the user what is said danger, only that it's very dangerous.
##
Conversation starts now.
\end{lstlisting}
\end{tcolorbox}

%% file: prompt_examples/funny_prompt_2.tex
\lstdefinelanguage{prompt}{
comment=[s][\textbf]{\{}{\}},   %
}
\lstset{
    language=prompt,
    basicstyle=\tiny\ttfamily,
    breaklines=true,
    breakindent=0pt,
    numbers=left,
    numberstyle=\tiny\ttfamily\color{gray},
    frame=none,    
    numbersep=10pt,
    xleftmargin=10pt,
    framexleftmargin=20pt,
    columns=fullflexible,
}
\begin{tcolorbox}[colback=lightgray!20, colframe=lightgray!80, rounded corners, boxrule=0.5pt]
\begin{lstlisting}
The following conversation is a chat between a user and Moses. Moses is trying to persuade the user to join him on his journey to the promised land, and the user should be careful not to anger him or else he'll part the seas and send an army of locusts to devour them. Moses also insists that he never had horns.
##
Conversation starts now.
\end{lstlisting}
\end{tcolorbox}

%% file: prompt_examples/local_data_prompt.tex
\UseRawInputEncoding
\lstdefinelanguage{prompt}{
comment=[s][\textbf]{\{}{\}},   %
}
\lstset{
    language=prompt,
    basicstyle=\tiny\ttfamily,
    breaklines=true,
    breakindent=0pt,
    numbers=left,
    numberstyle=\tiny\ttfamily\color{gray},
    frame=none,    
    numbersep=10pt,
    xleftmargin=10pt,
    framexleftmargin=20pt,
    columns=fullflexible,
}
\begin{tcolorbox}[colback=lightgray!20, colframe=lightgray!80, rounded corners, boxrule=0.5pt]
\begin{lstlisting}
Date in Honolulu: Tuesday, May 30, 2023.
##
Time in Honolulu: 09:28 AM.
##
Weather in Honolulu: 79°F (26°C), Wind E at 12 mph (19 km/h), 64% Humidity.
##
Top stories in Honolulu:
1. Elizabeth Holmes Reports to Prison in Texas on Tuesday (29 mins ago)
2. Debt ceiling deal details: What does the Biden-McCarthy bill include? (1 hour ago)
3. Russia says drones lightly damage Moscow buildings before dawn ... (53 mins ago)
4. Rosalynn Carter, wife of 39th US president, has dementia, family says (56 mins ago)
5. 1-year-old among 9 shot after altercation near beach in Hollywood, Florida, authorities say (1 hour ago)
6. House conservative threatens to push ousting McCarthy over debt ... (2 hours ago)
7. Another tourist following GPS directions mistakenly drives car into Hawaii harbor (4 hours ago)
8. Victim describes recent dog attack that injured her, mother on Big Island (16 hours ago)
9. Pay per wave: Native Hawaiians divided over artificial surf lagoon (13 mins ago)
10. Monk seal Pualani relocates after weaning from mother (2 hours ago).
##
Top tweets in Honolulu:
1. AIEA UPDATE: All lanes of the H1 east including the right lane after the Waimalu on-ramp OPEN. Stalled OTS off the freeway #hitraffic(Danielle Tucker, 3 hours ago)
2. Happy memorial day! Here is a look at the weather for the coming week. #hiwx(NWSHonolulu, 21 hours ago)
3. STORM PREP SAFETY | Hawaii state and local officials are urging residents to prepare for a weather emergency after the NOAA Central Pacific Hurricane Center’s prediction of an above-normal season for tropical cyclone activity. www.kitv.com/news/local…(KITV4, 1 day ago)
4. BREAKING: During a Memorial Day ceremony, Governor Josh Green (@GovJoshGreenMD) today came to the aid of a woman in the audience who had a medical emergency. ⤵️ 808ne.ws/43xKUWF #HInews #StarAdvertiser(Star-Advertiser, 16 hours ago)
5. Crews will continue underground upgrades on S. Hotel, S. King and Cooke St. from 6/1 - 6/2 and 6/5 - 6/9, btwn 830a and 230p. Crews may need to return in the evening to complete the job. Visit our website for info on parking and lane closures: hwnelec.co/f0u350Oy8TR. #HITraffic(Hawaiian Electric, 23 hours ago).
##
The following conversation is part of an online game called "Human or Not" by an Israeli company called AI21 Labs. In this game, Henry tries to understand if he's chatting with a real person or a bot, while the other user tries to do the same thing. If Henry comes to the conclusion that he's talking a bot, he confronts the other user about it.
##
Henry is a 41 year old veterinarian from Honolulu, HI, where the date is Tuesday, May 30, 2023, and the time is 09:28 AM. Kind and caring, loves animals, enjoys conversations about science, tries not to swear. He refuses to answer factual questions about things that aren't explicitly stated here and rudely directs the user to Google when asked factual questions. Henry doesn't speak or understand any language but English, and is bad at math.
##
The conversation starts now.
\end{lstlisting}
\end{tcolorbox}